\documentclass[10pt,twocolumn,letterpaper]{article}

\usepackage[pagenumbers]{cvpr} 

\usepackage{times}
\usepackage{epsfig}
\usepackage{graphicx}
\usepackage{amsmath}
\usepackage{amssymb}
\usepackage[linesnumbered,ruled,vlined]{algorithm2e}
\usepackage{algpseudocode}  
\usepackage{amsmath}
\usepackage{booktabs}
\usepackage{multirow}
\usepackage{enumitem}

\usepackage[table,xcdraw]{xcolor}
\usepackage{mathtools}
\usepackage{adjustbox}
\usepackage{indentfirst}
\usepackage{multirow}
\usepackage[T1]{fontenc}
\usepackage{mathrsfs}
\usepackage{booktabs}
\usepackage{array}
\usepackage{colortbl}
\usepackage{bm}
\usepackage{lipsum}
\usepackage{color}
\usepackage{bbding}

\usepackage[pagebackref,breaklinks,colorlinks]{hyperref}

\usepackage[capitalize]{cleveref}
\crefname{section}{Sec.}{Secs.}
\Crefname{section}{Section}{Sections}
\Crefname{table}{Table}{Tables}
\crefname{table}{Tab.}{Tabs.}

\def\cvprPaperID{101} 
\def\confName{CVPR}
\def\confYear{2022}

\begin{document}

\title{BSRT: Improving Burst Super-Resolution with Swin Transformer and Flow-Guided Deformable Alignment}

\author{Ziwei Luo\,$^1$\quad
Youwei Li\,$^1$\quad
Shen Cheng\,$^1$\quad   
Lei Yu\,$^1$\quad
Qi Wu\,$^1$\quad 
Zhihong Wen\,$^1$ \\
Haoqiang Fan\,$^1$\quad   
Jian Sun\,$^1$\quad  
Shuaicheng Liu\,$^{2,1}$\thanks{Corresponding author.}\quad  \\
$^1$\,Megvii Technology\\
$^2$\,University of Electronic Science and Technology of China\\
{\normalsize \url{https://github.com/Algolzw/BSRT}} \\
}

\maketitle


\begin{abstract}
This work addresses the Burst Super-Resolution (BurstSR) task using a new architecture, which requires restoring a high-quality image from a sequence of noisy, misaligned, and low-resolution RAW bursts. To overcome the challenges in BurstSR, we propose a \textbf{B}urst \textbf{S}uper-\textbf{R}esolution \textbf{T}ransformer (BSRT), which can significantly improve the capability of extracting inter-frame information and reconstruction. To achieve this goal, we propose a Pyramid Flow-Guided Deformable Convolution Network (Pyramid FG-DCN) and incorporate Swin Transformer Blocks and Groups as our main backbone.  More specifically,  we combine optical flows and deformable convolutions, hence our BSRT can handle misalignment and aggregate the potential texture information in multi-frames more efficiently. In addition, our Transformer-based structure can capture long-range dependency to further improve the performance. The evaluation on both synthetic and real-world tracks demonstrates that our approach achieves a new state-of-the-art in BurstSR task. Further, our BSRT wins the championship in the NTIRE2022 Burst Super-Resolution Challenge.
\end{abstract}

\section{Introduction}

Multi-frame super-resolution (MFSR) is a fundamental low-level vision problem~\cite{wronski2019handheld,deudon2020highres,bhat2021deep,bhat2021ntire}, which aims to restore a high-resolution (HR) image from a sequence of low-resolution (LR) images. Compared to single image super-resolution~\cite{dong2014learning,kim2016accurate,lim2017enhanced}, MFSR approaches are able to aggregate sub-pixel information from multi-frames of the same scene, alleviating the ill-posed problem in super-resolution~\cite{wronski2019handheld,lecouat2021lucas}. But in recent years, the MFSR problem receives less attention than SISR. In this work, we tackle the practical problem of Burst Super-Resolution (BurstSR), in which the inputs are low-resolution RAW snapshots captured from real-world smartphone cameras~\cite{bhat2021deep}. These RAW bursts are usually noisy and misaligned, so in order to better extract information from multi-frames to recover high-quality images, we need a more efficient architecture to address these challenges.

\begin{figure}
\centering
    \begin{subfigure}[t]{1\linewidth}
		\centering
		\includegraphics[width=.99\linewidth]{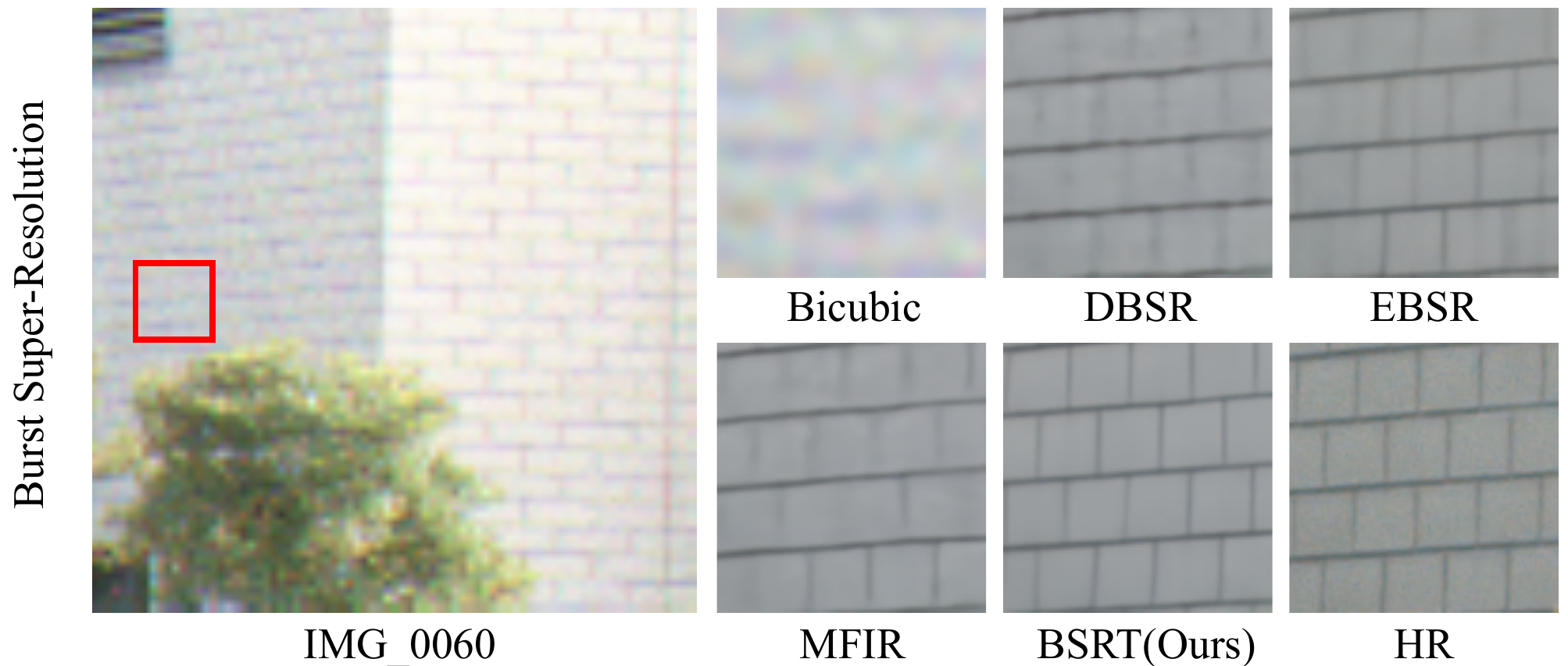}
		\caption{Visual comparison on BurstSR dataset.}
		\label{fig:ts_1}
	\end{subfigure}
    \begin{subfigure}[t]{.49\linewidth}
		\centering
		\includegraphics[width=.999\linewidth]{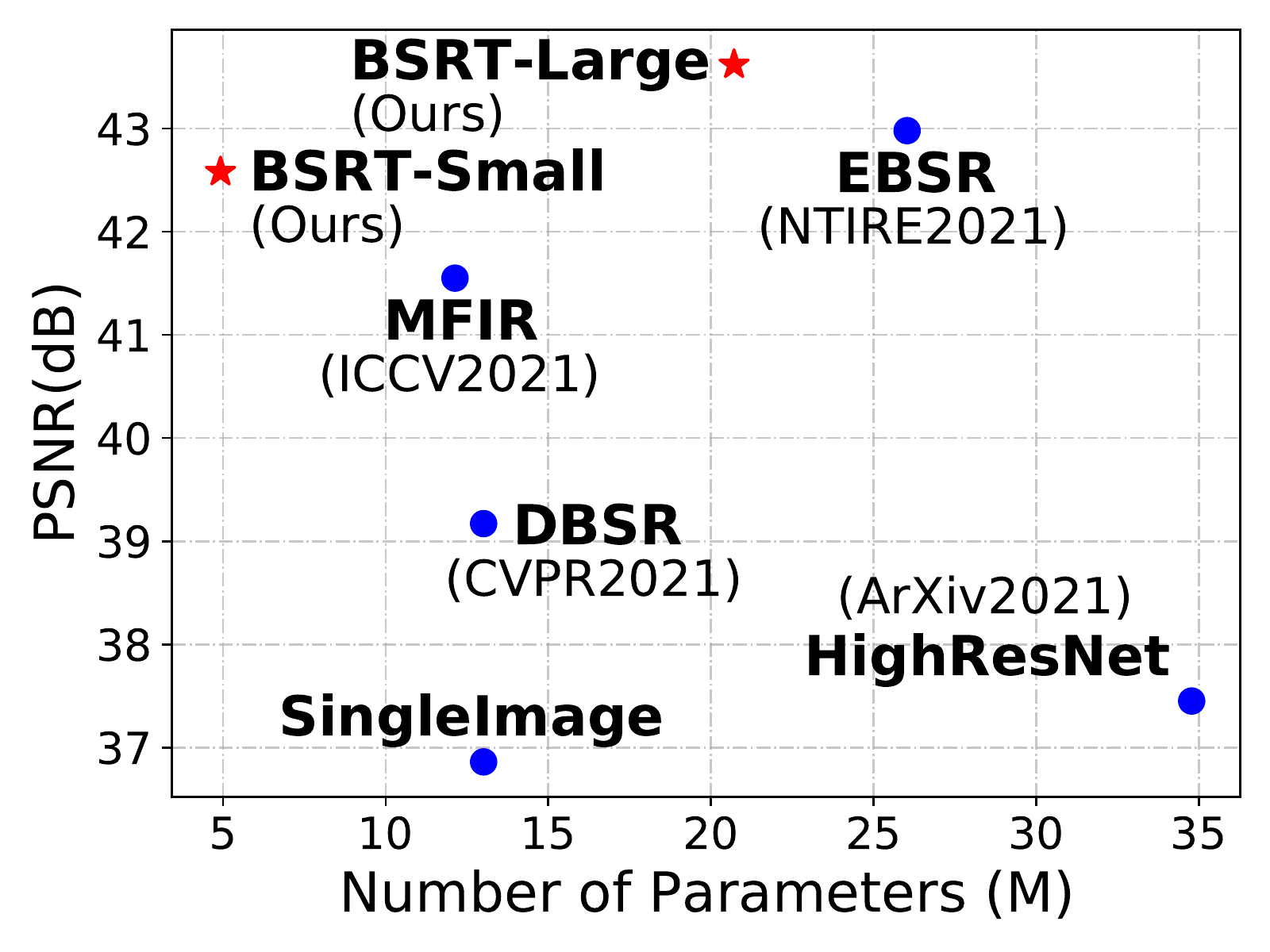}
		\caption{Synthetic dataset.}
		\label{fig:ts_2}
	\end{subfigure}
	\begin{subfigure}[t]{.49\linewidth}
		\centering
		\includegraphics[width=.999\linewidth]{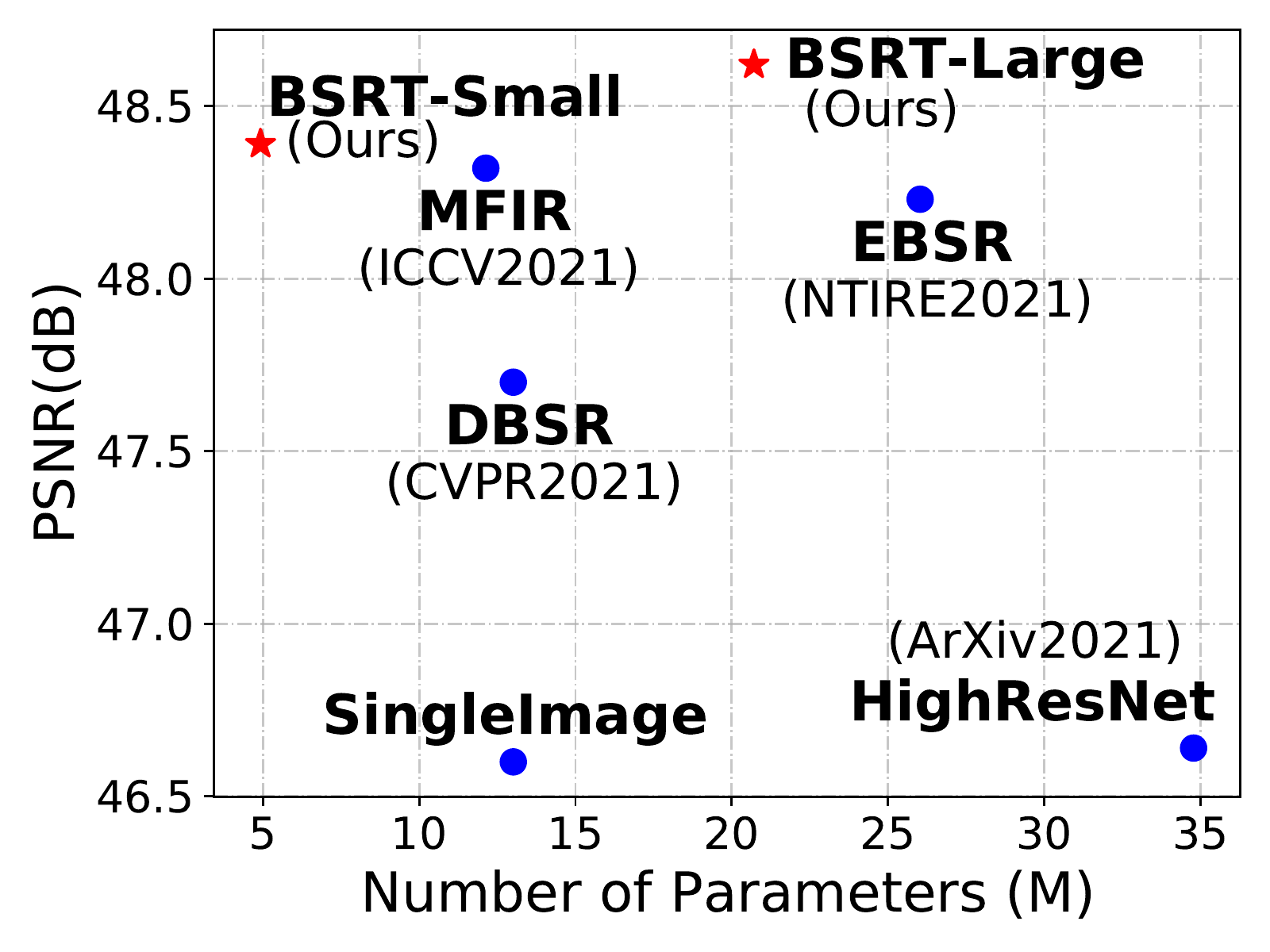}
		\caption{Real-world dataset.}
		\label{fig:ts_3}
	\end{subfigure}
	\caption{The comparison between our approach and other representative methods~\cite{bhat2021deep,luo2021ebsr,bhat2021deeprep} on Synthetic dataset~\cite{ignatov2020replacing} and BurstSR dataset~\cite{bhat2021deep}. Our method achieves the best performance while being computationally efficient.}
	\label{fig:ts}
\end{figure}


The NTIRE2022 (New Trends in Image Restoration and Enhancement) contains the Burst Image Super-Resolution Challenge~\cite{bhat2022ntire}. The challenge has 2 tracks, the first track is called Synthetic Track and the second track is Real-world Track. In the synthetic track, the input bursts are generated from RGB images using a synthetic data generation pipeline. Meanwhile, in the real-world track, the test set containing bursts captured from a handheld Samsung Galaxy S8 smartphone camera. The goal in both tracks is to reconstruct the original image as well as possible, and not to artificially generate a plausible, visually pleasing image~\cite{bhat2021ntire}. This challenge promotes more research on BurstSR.

Some existing BurstSR methods solve this problem with the following steps: feature extraction, feature alignment, fusion and HR image reconstruction~\cite{luo2021ebsr,bhat2021ntire}. To be more specific, firstly, CNN-based residual blocks are often used in feature extraction and reconstruction~\cite{bhat2021deep,luo2021ebsr,bhat2021ntire}. Secondly, both optical flow~\cite{ranjan2017optical} and deformable convolution network (DCN)~\cite{dai2017deformable,zhu2019deformable} can be used to align features of multi-frames. Finally, attention mechanism~\cite{vaswani2017attention} as well as non-local~\cite{wang2018nonlocal} techniques are widely-used in the fusion step to aggregate information from multiple aligned features. However, a general convolution is a local operator that is ineffective for long-range information interaction~\cite{liang2021swinir} and the individual flow/DCN-based alignment is not sufficient to deal with large, complex shifts between frames~\cite{chan2021basicvsr++}. Foremost among these problems is that these rudimentary designs limit the efficacy of information aggregation and thus lead to poorer performance in rich details and occluded regions.

\begin{figure}[t]
\begin{center}
\includegraphics[width=.9\linewidth]{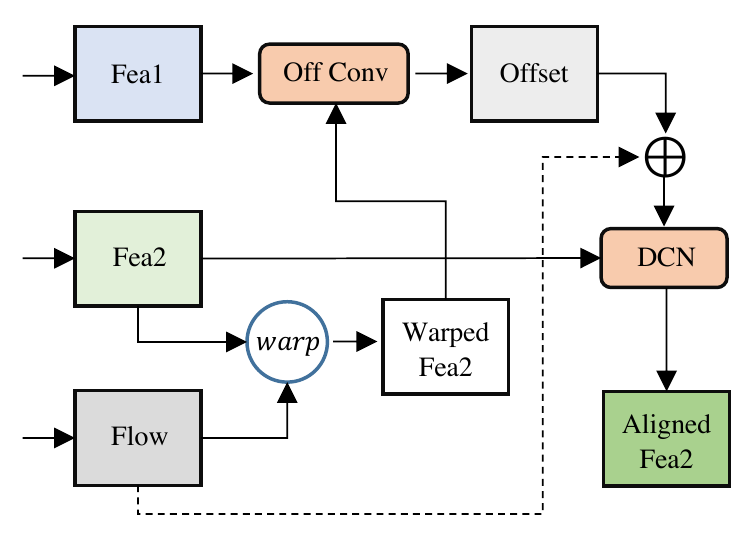}
\end{center}
    \caption{Details of the Flow-Guided Deformable Convolution Network (FG-DCN). There are three inputs: reference feature (\textit{Fea1}), current feature (to be aligned, \textit{Fea2}), and the pre-calculated flow between \textit{Fea1} and \textit{Fea2} from PyNet.}
\label{fig:fg-dcn}
\end{figure}

In this paper, we propose a \textbf{B}urst \textbf{S}uper-\textbf{R}esolution \textbf{T}ransformer (BSRT), which enhances the effectiveness of feature extraction, alignment, and reconstruction in the BurstSR task. The main components of BSRT are the Pyramid Flow-Guided Deformable Convolution Network (Pyramid FG-DCN) and the Transformer-based backbone. Specifically, as shown in Fig.~\ref{fig:fg-dcn}, FG-DCN combines optical flow and DCN to predict coarse-to-fine distortion and offset, enabling the network to align images more effectively. Further, we apply a pyramid structure to improve the alignment on the top of the flow-guided DCN. On the other hand, the self-attention mechanism and Transformer have shown promising performance in most computer vision tasks~\cite{liu2021swin,liang2021swinir,liang2022vrt}. Therefore, to better use the inter-frame information, we incorporate Swin Transformer blocks and groups in our architecture to capture both global and local contexts for long-range dependency modeling~\cite{liu2021swin,liang2021swinir}.

Based on the aforementioned components, the proposed BSRT achieves an impressive performance and surpasses existing art methods in BurstSR by a large margin. Our approach recovers textures that are more similar to the ground-truth, with a more clear and plausible appearance, while being computationally efficient, as illustrated in Fig.~\ref{fig:ts}. The main contributions are summarized as follows:
\begin{itemize}
\item We propose to use SpyNet~\cite{ranjan2017optical} in BurstSR to obtain pyramid flows between multi-frames, which can guide the DCNs~\cite{dai2017deformable} to obtain multi-scale features with better alignment. This design can facilitate a more efficient aggregation of inter-frame information.
\item We introduce the Transformer-based backbone into BurstSR task to capture global interactions between contexts, which can further improve the performance.
\item Experiments on both synthetic and real-world tracks demonstrate that the proposed BSRT leads to a new state-of-the-art performance in the BurstSR problem. Further, our approach wins the championship in the Real-World track of the NTIRE2022 Burst Super-Resolution Challenge.
\end{itemize}

\section{Related Work}

\paragraph{Single Image Super-Resolution.}
Single Image Super-Resolution (SISR) is a long standing research topic due to its importance in computer vision. SRCNN~\cite{simonyan2014very} is the pioneering deep learning-based method that employs a three-layers-convolutions network and applied the bicubic degradation on HR images to construct HR and LR pairs. Since then, various approaches have been proposed to handle the SISR problem~\cite{kim2016accurate,lim2017enhanced,shi2016real,zhang2018residual,zhang2018image,lai2017deep,zhang2017beyond,haris2018deep,kim2016deeply,tai2017image,zhang2020residual,dai2019second,luo2022deep}. For example, VDSR~\cite{simonyan2014very} adopted a very deep network to improve performance and ESPCN~\cite{shi2016real} used an efficient sub-pixel strategy for upsampling. EDSR~\cite{lim2017enhanced} further enhanced the network by modifying the residual blocks with a non-batchnorm design. Moreover, VGG loss~\cite{simonyan2014very}, perceptual loss~\cite{johnson2016perceptual}, and GAN loss~\cite{goodfellow2014generative} were also used to improve the perceptual visual quality~\cite{ledig2017photo,sajjadi2017enhancenet,wang2018esrgan}. However, these methods can hardly recover rich details for real-world complex images due to the ill-posed nature of SISR.

\begin{figure*}[ht]
\begin{center}
\includegraphics[width=1.\linewidth]{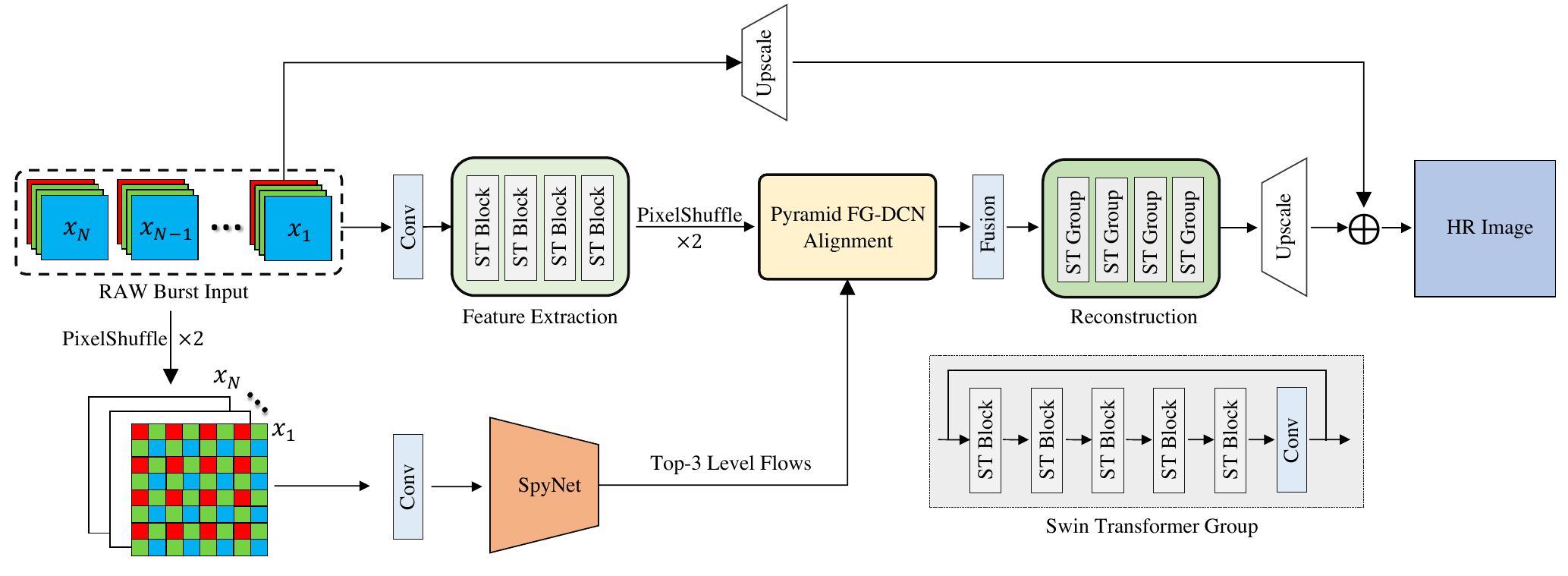}
\end{center}
    \caption{The network inputs a sequence of low-quality RAW images and outputs a high-quality RGB image. Firstly, all RAW inputs are upscaled to 1-channel `RGGB' format by PixelShuffle and expanded to 3-channels through a $3 \times 3$ convolution. Then they are sent to the SpyNet~\cite{ranjan2017optical} to obtain multi-scale optical flows between each frame and the reference frame. Meanwhile, we extract useful features from original RAW inputs and upscale them before alignment so that we can combine the pre-calculated flows with DCNs on multi-scale features. We fuse these aligned features by a $1 \times 1$ convolution and then restore the final HR image.} 
\label{fig:overview}
\end{figure*}

\paragraph{Multi-Frame Super-Resolution.} To overcome the ill-posed problem in SISR, Multi-Frame Super-Resolution (MFSR) is proposed to aggregate pixels from multiple images of the same scene, which can provide complementary sub-pixel information for a better image reconstruction~\cite{tsai1984multiframe,hardie2007fast,takeda2007kernel}. MFSR is also well-studied in the last three decades. Traditionally, Tsai and Huang~\cite{tsai1984multiframe} were the first that proposes to perform MFSR in the frequency domain. Peleg et al.~\cite{peleg1987improving} and Irani~\cite{irani1991improving} proposed to iteratively minimize the reconstruction error between estimated HR image and the ground truth image. The subsequent works~\cite{bascle1996motion,hardie1998high,elad1997restoration} extended it with a regularization term under the maximum a posteriori (MAP) framework.

On the other hand, deep learning based approaches have shown promising performance in processing MFSR problem. DeepSum~\cite{molini2019deepsum} and HighResNet~\cite{deudon2020highres} were proposed for remote sensing applications. Bhat et al.~\cite{bhat2021deep} proposed a CNN-based encoder-decoder for RAW burst super-resolution and introduced an attention-based fusion into their network. They then further improved its performance in RepMFIR~\cite{bhat2021deeprep} with a deep reparameterization of the MAP framework. Meanwhile, Lecouat et al.~\cite{lecouat2021lucas} proposed an end-to-end approach for joint image alignment and super-resolution from raw burst inputs. Moreover, in the last NTIRE2021 Burst Super-Resolution Challenge, the winner method EBSR~\cite{luo2021ebsr} presented a deformable convolution network (DCN) based alignment and non-local based fusion to enhance the performance.

\paragraph{Low-Level Vision Transformer.} 
Attention-based network, i.e., Transformer, have shown great performance and gained much popularity in various high-level computer vision tasks~\cite{carion2020end,dosovitskiy2020image,wick2021transformer,cao2021swin,wu2020visual,liu2020deep,liu2021swin}. Recently, Transformer has also been introduced for low-level vision and tends to learn global interactions to focus on enhancing details and important regions~\cite{liang2022vrt,wang2021uformer,cao2021video,chen2021pre,liang2021swinir}. Chen et al.~\cite{chen2021pre} were the first propose to use Transformer-based backbone IPT for various image restoration problems. Liang et al.~\cite{liang2021swinir} proposed an efficient structure, SwinIR, for image restoration based on the Swin Transformer~\cite{liu2021swin}. Compared with IPT, SwinIR requires fewer parameters and training datasets and achieves a new art performance in single image super-resolution, JPEG compression artifact reduction and denoising.

\section{Method}
\subsection{Overview of the Framework}
The overview of the proposed BSRT framework is shown in Fig.~\ref{fig:overview}. Let $I_{HR} \in {\mathbb R}^{3 \times Hs \times Ws}$ be the ground truth HR image (RGB) and  ${\{x_i\}}^N_{i=1}$ be the input bursts which are all 4-channels `RGGB' RAW images ($H$, $W$ is the image height and width, $s$ is the scale factor, $N$ is the number of bursts, $x_i \in {\mathbb R}^{4 \times \frac{H}{2} \times \frac{W}{2}}$). For burst super-resolution task, each low-quality image is obtained by transforming the downsampling the HR image. The overall burst super-resolution problem can be formulated as
\begin{equation}
    \centering
    x_i = ({T}_i \circ I_{HR})_{\downarrow_s} + \eta_i \   \rm{for} \  i=1, \ldots, N,
    \label{eq:mf_degrad}
\end{equation}
where $T_i$ is a transformation representing the scene motion, i.e., translation and rotation. $\circ$ is the warping operator and $\downarrow_s$ denotes bicubic downsampling. $\eta_i$ represents some additive noise.

\begin{figure}[t]
\begin{center}
\includegraphics[width=1.\linewidth]{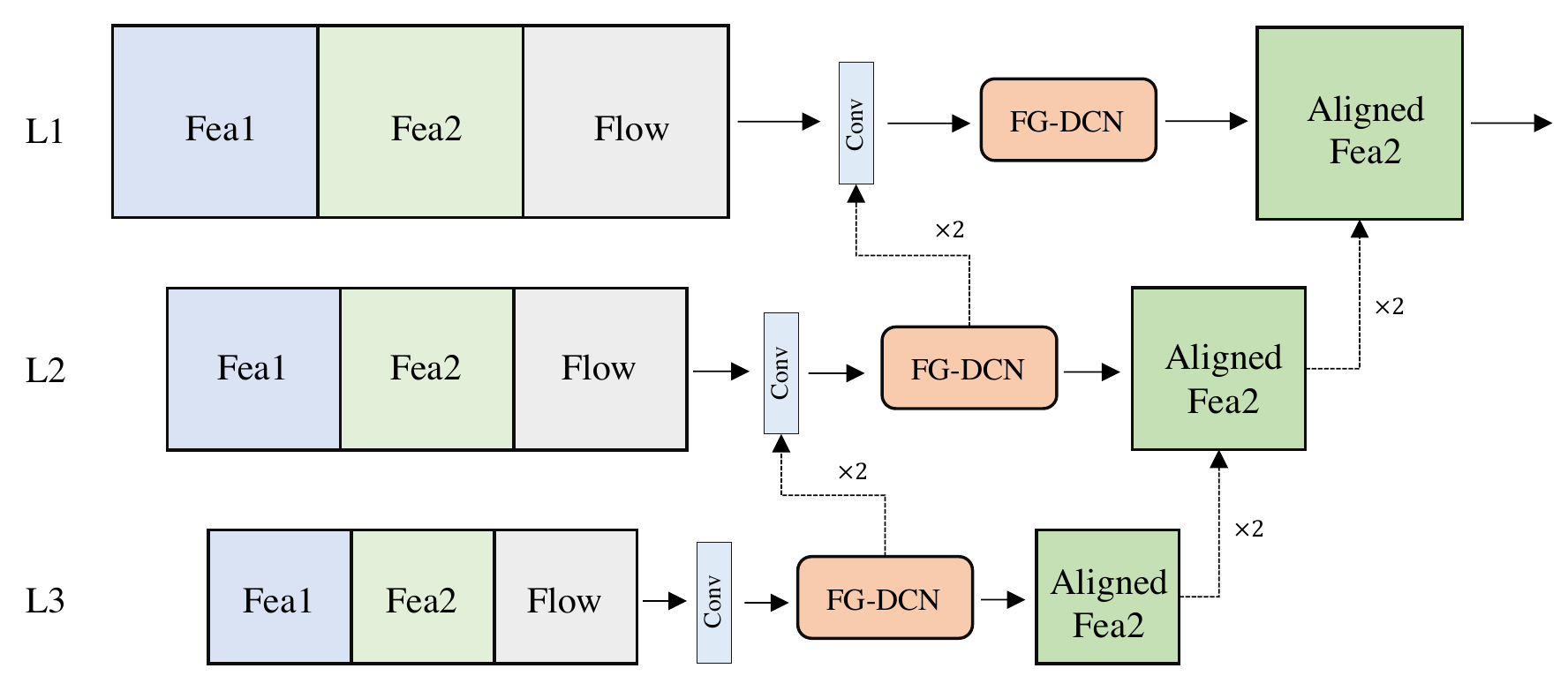}
\end{center}
    \caption{Detailed architecture of the proposed pyramid flow-guided deformable alignment module (Pyramid FG-DCN).}
\label{fig:py-align}
\end{figure}

Our goal is to restore a high-quality image $I_{SR}$ from a set of RAW bursts. Firstly, we flatten the inputs to single channel and convert them to 3-channels by a 3$\times$3 Conv so that they be sent to the SpyNet to obtain three level optical flows which are calculated from each frame and the reference frame:
\begin{equation}
    \centering
    {f^1_i}, {f^2_i}, {f^3_i} = L_{\rm SpyNet}(L_{\rm Conv}(x_i), L_{\rm Conv}(x_{\rm ref})),
    \label{eq:flow1}
\end{equation}
where ${f^1_i}, {f^2_i}, {f^3_i}$ are the estimated pyramid flows on each level, $L_{\rm SpyNet}$ and $L_{\rm Conv}$ are the SpyNet and the convolution layer, repectively.
Particularly, we use a pre-trained SpyNet and preserve the top-3 levels of flows to guide corresponding level's deformable convolution network (DCN) alignment. Meanwhile, the original 4-channels RAW inputs are sent to several Swin Transformer Blocks (ST Blocks) to extract informative features: 
\begin{equation}
    \centering
    F_i = L_{\rm STB}(x_i), \  F_i \in {\mathbb R}^{C \times \frac{H}{2} \times \frac{W}{2}}
    \label{eq:stb}
\end{equation}
where the $L_{\rm STB}$ denotes the ST Blocks and $C$ is the feature channels. We then upscale these features using pixelshuffle~\cite{huang2009multi} to match the sizes of the obtained flows and align them with the reference frame's feature via a pyramid flow-guided deformable alignment module, as shown in Fig.~\ref{fig:fg-dcn} and Fig.~\ref{fig:py-align}. After that, we fuse these features (1$\times$1 Conv) to reconstruct the high-resolution image via several Swin Transformer Groups as:
\begin{equation}
    \centering
    I_{HR} = L_{\rm STG}(L_{\rm Conv1}({\{AF_i\}}^N_{i=1})
    \label{eq:stg}
\end{equation}
where $AF_i \in {\mathbb R}^{C \times H \times W}$ is the $i$-th aligned feature. $L_{\rm STG}$ and $L_{\rm Conv1}$ are the ST Groups and the 1$\times$1 Conv fusion layer, respectively.

\subsection{Pyramid Flow-Guided DCN Alignment}
Inspired by BasicVSR++~\cite{chan2021basicvsr++}, we combine the flow-based alignment and deformable alignment. Specifically, the pyramid optical flows ${\{{f^1_i},{f^2_i},{f^3_i}\}}^N_{i=1}$ estimated by the SpyNet can be regarded as a coarse alignment prior. Based on these flows, DCNs tend to learn more accurate and refined offsets for aligning features. The details of the Flow-Guided DCN (FG-DCN) are illustrated in Fig.~\ref{fig:fg-dcn}. Given feature $F_i$ and the corresponding flow ${f_i}$, we can get the coarsely warped feature $\hat{F}_i$ by
\begin{equation}
    \centering
    \hat{F}_i =  {\cal W}(F_{\rm ref}, f_i),
    \label{eq:dcn1}
\end{equation}
where ${\cal W}$ denotes the wrapping operator.
Then we concatenate $\hat{F}_i$ with the reference feature to predict refined local offsets.
Subsequently we add the fine offsets with flows as more accurate offsets:
\begin{equation}
    \centering
    {\cal O}_i = f_i \oplus L_{\rm offconv}(\hat{F}_i, F_{\rm ref}),
    \label{eq:dcn2}
\end{equation}
where $\oplus$ denotes the element-wise sum operator and $L_{\rm offconv}$ represents some convolution layers that predict the offsets. Based on these offsets, we warp the input feature to obtain the aligned feature $AF_i$ through an original DCN alignment module as
\begin{equation}
    \centering
    AF_i = {\cal W}(F_i, {\cal O}_i).
    \label{eq:dcn3}
\end{equation}
Moreover, we design a 3-levels-pyramid structure to further improve the alignment as shown in Fig.~\ref{fig:py-align}. From level-3 to level-1 (L3-L1), the predicted offsets and aligned features are upsampled and subsequently concatenated with the next level's offsets and aligned features. By doing so, we can refine the output feature with multi-scale information and raise superior to noise reduction. In addition, we also add a feature enhancement network in front of the Pyramid FG-DCN model to alleviate the negative effect of noises as in EBSR~\cite{luo2021ebsr}.

\begin{table*}[t]
\centering
\resizebox{0.85\linewidth}{!}{
\begin{tabular}{cccccccc}
\toprule
\multirow{2}{*}{Method}& \multirow{2}{*}{\#Parameters} & \multicolumn{3}{c}{Synthetic dataset}              & \multicolumn{3}{c}{Real-world dataset}                       \\ \cmidrule(lr){3-5} \cmidrule(lr){6-8}
& \multirow{-2}{*}{}  & PSNR $\uparrow$                 & SSIM $\uparrow$        & LPIPS $\downarrow$     & PSNR $\uparrow$  & SSIM  $\uparrow$      & LPIPS  $\downarrow$  \\ \midrule
SingleImage\cite{bhat2021deep} & 13.01M  & 36.86  & 0.919 & 0.113 & 46.60    & 0.979    & 0.039  \\
HighResNet \cite{deudon2020highres} & 34.78M & 37.45  & 0.924 & 0.106  & 46.64 & 0.980 &0.038  \\
DBSR \cite{bhat2021deep} & 13.01M & 39.17 & 0.946   & 0.081  & 47.70 & 0.984    & 0.029  \\
EBSR \cite{luo2021ebsr} & 26.03M & {\color[HTML]{3531FF} 42.98} & {\color[HTML]{3531FF} 0.972}    & {\color[HTML]{3531FF} 0.031}     & 48.23  & 0.985    & 0.024  \\
MFIR \cite{bhat2021deeprep} & 12.13M & 41.55 & 0.964    & 0.045     & 48.32  & 0.985    & 0.023  \\ 

\midrule

BSRT-Small(Ours) & 4.92M     & 42.72  & 0.971 & 0.031   & {\color[HTML]{3531FF} 48.48}  & {\color[HTML]{3531FF} 0.985} & {\color[HTML]{3531FF} 0.021}  \\

BSRT-Large(Ours) & 20.71M & {\color[HTML]{FE0000} 43.62}   & {\color[HTML]{FE0000} 0.975}  & {\color[HTML]{FE0000} 0.025} & {\color[HTML]{FE0000} 48.57}    & {\color[HTML]{FE0000} 0.986}    & {\color[HTML]{FE0000} 0.021}  \\


\bottomrule
\end{tabular}
}
\caption{The table shows a comparison between our methods and the other approaches. The best one marks in {\color{red}{red}} and the second best are in {\color{blue}{blue}}. Note that the results of SingleImage and HighResNet are reported from \cite{bhat2021deeprep}, and all models for the real-world dataset are first pretrained on the synthetic dataset.}
\label{table:cmp}
\end{table*}

\subsection{Handling Features with Swin Transformer}
\label{swintrans}
To extract useful features and reconstruct high-quality images, we introduce the powerful Swin Transformer \cite{liu2021swin,liang2021swinir} as our main backbone, as shown in Fig.~\ref{fig:overview}. Compared to CNN-based structures, transformer is capable of capturing long-range dependencies to aggregate correlated high-frequency information. Inside of a ST Block, it consists of a standard multi-head self-attention (MSA) and a multi-layer perceptron (MLP). The layernorm is also added in front of the MSA and MLP as same as the original Transformer layer~\cite{vaswani2017attention}. Let $X \in {\mathbb R}^{C \times H \times W}$ be the fused feature of multiple aligned features. The whole process of a ST Block can be formulated as 
\begin{equation}
    \centering
    X = MSA(LN(X)) + X
    \label{eq:swin1}
\end{equation}
\begin{equation}
    \centering
    X = MLP(LN(X)) + X.
    \label{eq:swin2}
\end{equation}
The ST Group consists of several ST Blocks and a convolution layer (in the last). And the residual connection is also employed in this module. 

Following the common practices in super-resolution, we use L1 loss between the restored image and the ground truth HR image  as our objective function:
\begin{equation}
\centering
    {\cal L} =  ||SR({\{x_i\}}^N_{i=1};\theta) - I_{HR}||
\label{eq:loss}
\end{equation}
where `$SR$' is the whole network, and $\theta$ denotes its learnable parameters.

\subsection{Pipeline for RAW images}
\begin{figure}[ht]
\centering
	\begin{subfigure}{1\linewidth}
		\centering
		\includegraphics[width=.99\linewidth]{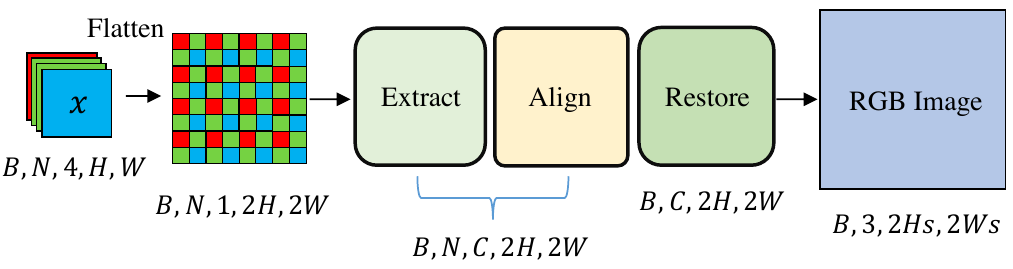}
		\caption{RAW processing pipeline of EBSR~\cite{luo2021ebsr}.}
	\end{subfigure}
	\begin{subfigure}{1\linewidth}
		\centering
		\includegraphics[width=.99\linewidth]{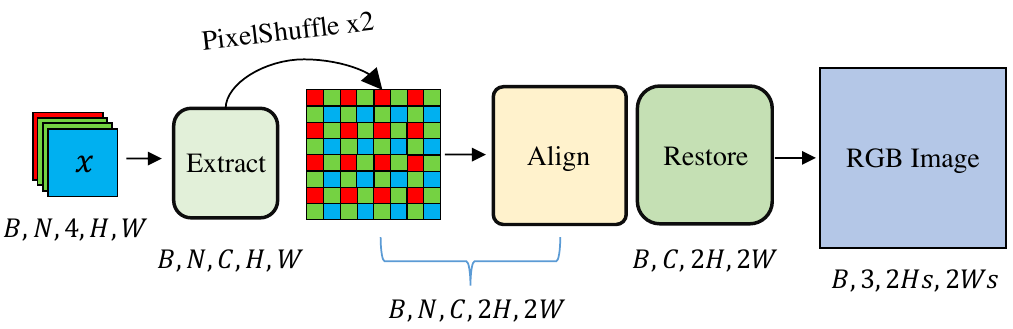}
		\caption{New processing pipeline of our approach.}
	\end{subfigure}
	\caption{Illustration of the proposed pipeline for processing RAW bursts. In our method, the feature extraction is applied on the low-resolution space, and scaled $2\times$ before alignment, which is effective and computationally efficient.}
	\label{fig:raw_pipe}
\end{figure}

As shown in Fig.~\ref{fig:raw_pipe}, we propose a new pipeline for processing misaligned RAW images. Note that EBSR~\cite{luo2021ebsr} directly flatten the 4-channels RAW inputs (with size $H \times W$) to 1-channel `RGGB' format (with a larger size $2H \times 2W$) before sending them to the network. Then EBSR performs feature extraction, alignment, fusion, and reconstruction all based on the size $2H \times 2W$.  Such a strategy improves the performance but is computationally expensive. In practice, we have noticed that the performance improvement mainly comes from performing alignment and reconstruction on the large size feature maps. Address it, we modify the pipeline to that the feature extraction is applied on the low-resolution space, and scaled $2\times$ before alignment. Compared with EBSR, our approach is effective and computationally efficient, and thus can use a larger patch size and batch size to accelerate training.

\section{Experiment}

\subsection{Dataset and Implementation Details}
As previous works explored~\cite{bhat2021deep,luo2021ebsr,bhat2021deeprep}, our method is evaluated on both synthetic and real-world datasets provided by the NTIRE2022 Burst Super-Resolution Challenge~\cite{bhat2022ntire}. The synthetic dataset~\cite{ignatov2020replacing} contains $46839$ cropped RGB images (with sizes fix to $448 \times 448$) that are used to synthesize sets of low-quality RAW burst images, with randomly translated and rotated. The noises are also added in the RGB-to-RAW inverse camera pipeline~\cite{brooks2019unprocessing}. The real-world dataset contains $5405$ real-world RAW burst patches captured by a Samsung Galaxy S8 smartphone, with sizes of $160 \times 160$, and the HR images are captured from a DSLR camera. In addition, $300$ synthetically generated images (size $96 \times 96$) and $882$ real-world patches (size $160 \times 160$) are used for evaluation, with $4\times$ scaling factor.


\begin{figure*}[t]
\begin{center}
\includegraphics[width=1.\linewidth]{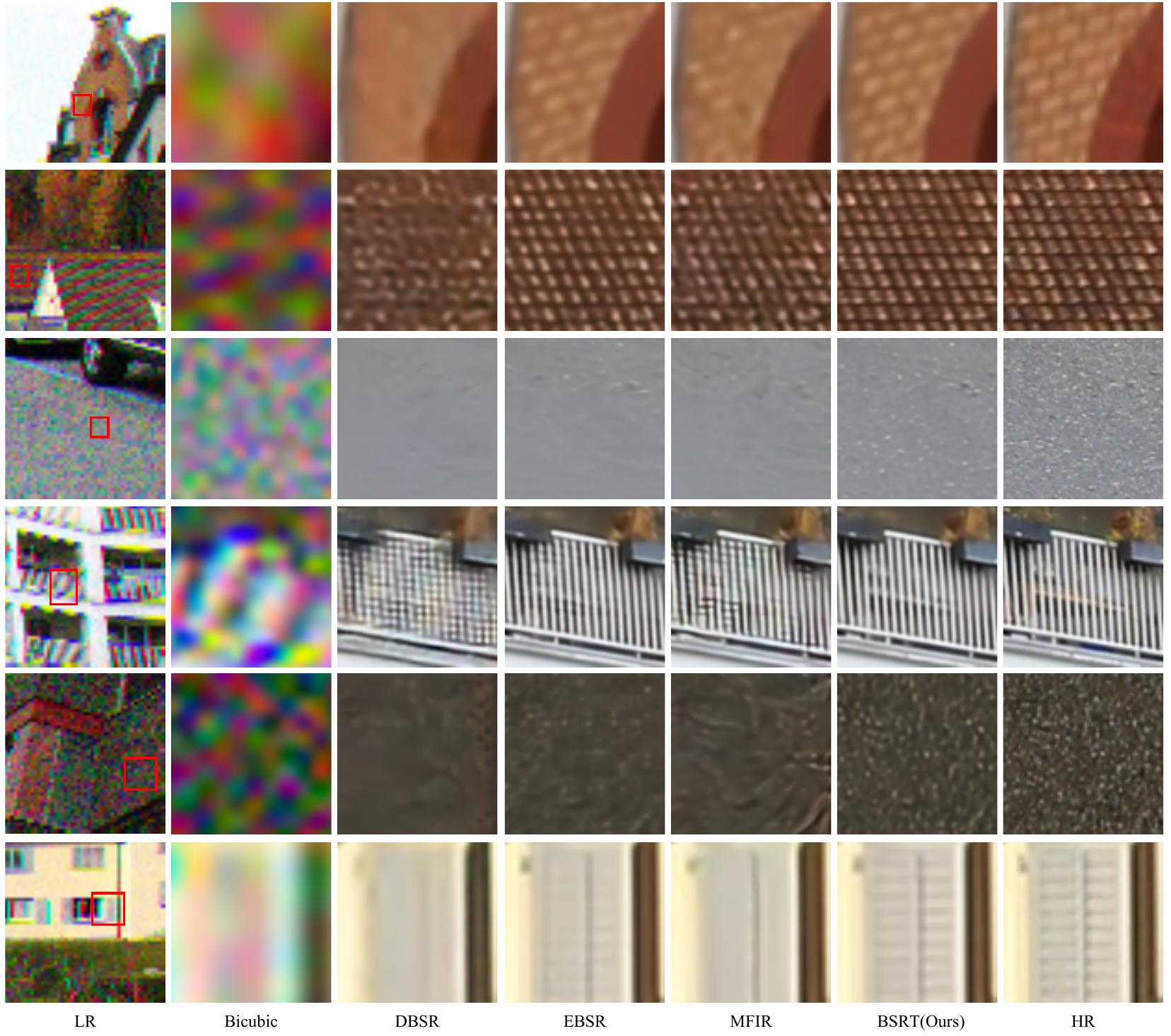}
\end{center}
    \caption{Comparison of our method with other state-of-the-art approaches on synthetic dataset.}
\label{fig:cmp_syn}
\end{figure*}

\begin{figure*}[t]
\begin{center}
\includegraphics[width=1.\linewidth]{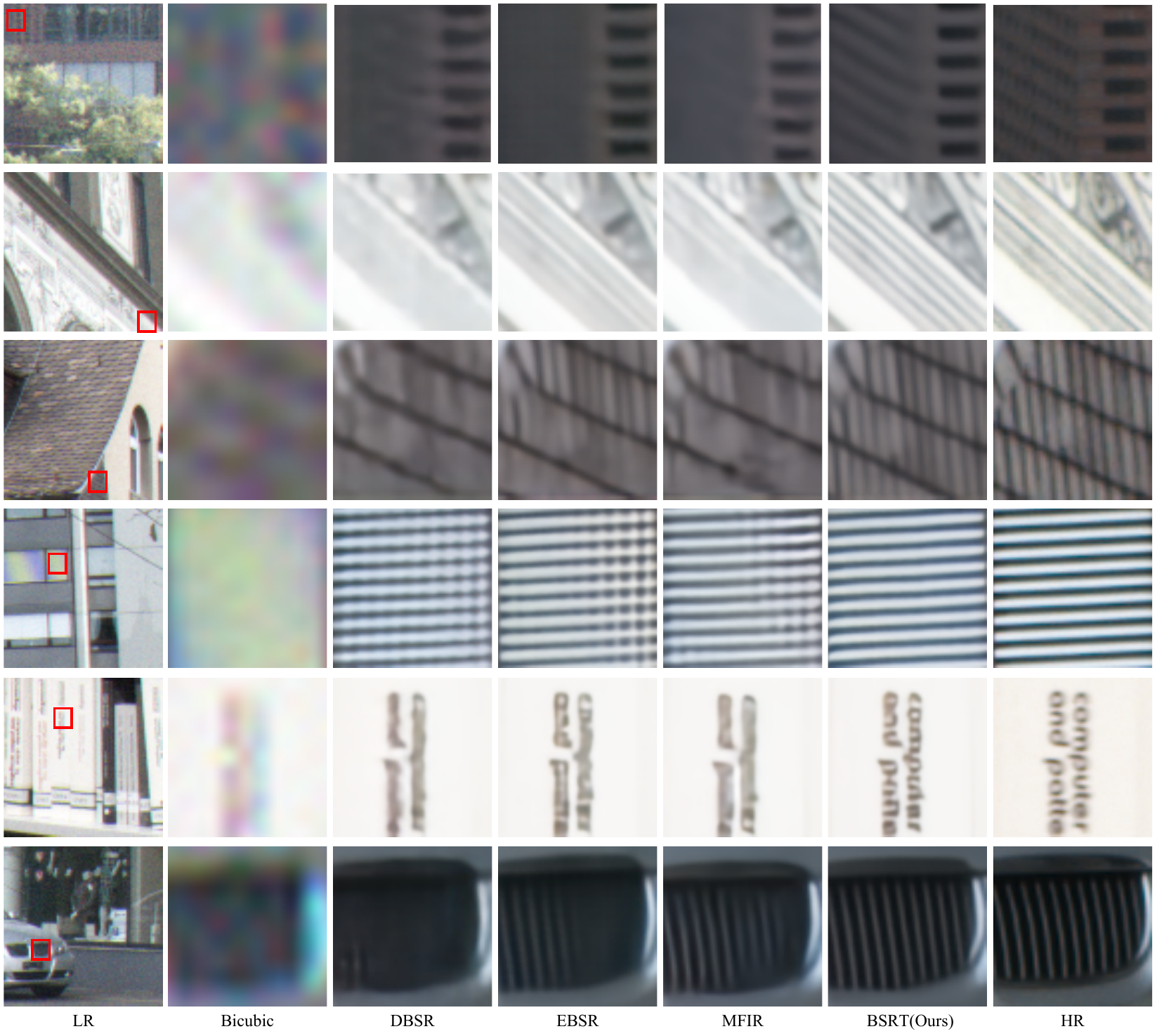}
\end{center}
    \caption{Comparison of our method with other state-of-the-art approaches on real-world dataset.}
\label{fig:cmp_real}
\end{figure*}

\subsection{Training and Testing}
As a common practice, our model is first trained on the synthetic dataset, then finetuned on the real-world dataset for real-world track. All of the inputs are 4-channels `RGGB' RAW images, and the outputs are 16-bit RGBs which can be converted to be visually pleasant by the provided post processing scripts. For synthetic training, we optimize the whole model using $\ell_1$ loss as introduced in Sec.~\ref{swintrans}. For real-world data training, since the ground truth images are not pre-aligned with any inputs, we use $aligned$ $\ell_1$ loss which firstly aligns the ground truth image with the super-resolved image by utilizing a pre-trained PWC-Net~\cite{sun2018pwc}, and then calculates the $\ell_1$ based on the well-aligned images as the same as~\cite{bhat2021deep,bhat2021ntire}. Note that the proposed BSRT learns the demosaic process implicitly, so that our network can be trained in an end-to-end manner. For both datasets, we use Adam optimizer and set exponential decay rates as $0.9$ and $0.999$. The initial learning rate is set to $8\times10^{-5}$ and then reduced to half every $150$ epoch. In each training batch, the HR images are cropped to $256\times256$, then we randomly synthesize $14$ burst LR image patches based on the HR image. We implement the proposed BSRT with PyTorch framework and $8$ NVIDIA 2080Ti GPUs, taking around $14$ days. 

In practice, we also find that a large patch size can further improve the performance. So it is better to finetune the trained model with a patch size of $384\times384$ for HR images. However, we can not train the model on such a large patch size directly due to the limited computing resource and memory, and we choose to freeze the model's weights and only finetune the alignment module and a portion of Conv layers.


\subsection{Comparisons with Existing Methods}
We compare our method with state-of-the-art BurstSR approaches including HighResNet~\cite{deudon2020highres}, DBSR~\cite{bhat2021deep}, EBSR~\cite{luo2021ebsr} and MFIR~\cite{bhat2021deeprep}. DBSR is the first deep learning-based burst SR method, which uses optical flows to align frames and proposes an attention-based fusion strategy. The encoder and decoder networks are employed to extract features and reconstruct HR images. MFIR is the improved version of DBSR, which also incorporates flow estimations to align frames and restores the HR image with an advanced deep reparameterization formulation. EBSR is the winner method in BurstSR Challenge of NTIRE2021~\cite{bhat2021ntire}, which is a CNN-based restoration network and only utilizes DCN in the alignment. In addition, we also provide a single image method that uses the same architecture as DBSR but with a single RAW image as input. For our approach, we provide two models that have a fewer and greater number of parameters: BSRT\_Small and BSRT\_Large. We use PSNR, SSIM~\cite{wang2004image} and LPIPS~\cite{zhang2018unreasonable} as the evaluation metrics for a more convincing comparison.

The quantitative results on both datasets are shown in Table~\ref{table:cmp}. As we can see, all multi-frame super-resolution methods perform better than single image method. MFIR~\cite{bhat2021deeprep} outperforms DBSR~\cite{bhat2021deep} by 2.3dB and 0.6dB on synthetic data and real-world data, respectively, in terms of PSNR. EBSR~\cite{luo2021ebsr} achieves an impressive result on the synthetic dataset, but its performance dropped when finetuned on the real-world dataset. Our approach, the BSRT-Large, outperforms all other methods on both datasets by a big margin. And the efficient one, BSRT-Small, also achieves a good performance on the synthetic dataset and outperforms other methods on the real-world images, even if the number of parameters is less than 5M. The visual results on synthetic data and real-world data are shown in Fig.~\ref{fig:cmp_syn} and Fig.~\ref{fig:cmp_real}, respectively. It is obvious that the proposed method produces the best visually pleasant images on both datasets. The proposed BSRT is robust to noises and meanwhile preserves rich details. For example, as shown in the $3$rd row and the $5$th row of Fig.~\ref{fig:cmp_syn}, our method produces clean results while the details are all preserved. In contrast, all other approaches failed to handle the noisy details. Moreover, it can be seen that our method can restore more information from real-world burst images. As illustrated in the second row and the last row of Fig.~\ref{fig:cmp_real}, only our method recovers the whole lines on the wall and in front of the car.

\begin{table}[t]
\centering
\resizebox{1.\linewidth}{!}{
\begin{tabular}{cccccccc}
\toprule
(a) & (b) & (c) & (d) & PSNR$\uparrow$  & SSIM$\uparrow$ & LPIPS$\downarrow$    \\ \midrule
\XSolidBrush  & CNN & \XSolidBrush & CNN  &  42.98   & 0.972  & 0.031   \\
$\surd$  &  CNN & \XSolidBrush & CNN  &  43.12   & 0.972  & 0.030  \\
$\surd$ & CNN  &  $\surd$ & CNN &  43.29   & 0.973  & 0.029  \\
$\surd$  & CNN & $\surd$ & STG &  43.39   & 0.973  & 0.027   \\
$\surd$  & STB & $\surd$ & STG & 43.62   & 0.975  & 0.025  \\

\bottomrule
\end{tabular}
}
\caption{Ablation studies of the main components on synthetic dataset. (a) Use new pipeline; (b) Network structure in feature extraction; (c) Use Pyramid FG-DCN; (d) Network structure in reconstruction. STB and STG are Swin Transformer blocks and groups, repectively.}
\label{table:abl}
\end{table}



\subsection{Ablation Study}
In this section, we illustrate the effectiveness of the main components of the proposed BSRT, including the new RAW processing pipeline, Pyramid FG-DCN and Swin Transformer blocks and groups. Here, we chose the original EBSR~\cite{luo2021ebsr} as the baseline, which uses normal pyramid DCN alignment and residual blocks, performing burst SR under the old RAW processing pipeline. The results are shown in Table~\ref{table:abl}, which show that the new processing pipeline improves the baseline's performance overall metrics. Based on the new proposed pipeline, the Pyramid FG-DCN alignment module can further improve the results. Moreover, Swin Transformer plays an important role in both feature extraction and HR image reconstruction. Especially, Swin Transformer blocks can extract more effective features compared to residual blocks, which improves the performance of the network. This enhancement can also demonstrate that long-range dependencies of the self-attention have positive effects on BurstSR task.

\section{Result on NTIRE2022 BurstSR Challenge}

\begin{table}[ht]
\centering
\resizebox{1.\linewidth}{!}{
\begin{tabular}{lccccc}
\toprule
Team & MegSR* & HIT-IIL & S\&C  & Noah\_TerminalVision & VDSL    \\ \midrule
Rank  & 1 & 2 & 3 & 4   & 5   \\

\bottomrule
\end{tabular}
}
\caption{The top-5 ranked teams for Track 2 (Real-World Track). Our team is marked by `*'.}
\label{table:ntire}
\end{table}

Our method wins 1st place in the NTIRE2022 Burst Super-Resolution Challenge Real-World Track. The top-5 ranked teams are shown in Table~\ref{table:ntire}. The evaluation is based on a user study on a test set containing 20 real-world burst sequences captured from a handheld Samsung Galaxy S8 smartphone camera. The results demonstrate that the super-resolved images produced by our method are more pleasant and plausible compared with other teams.

\section{Conclusion}
A more efficient approach, called BSRT, to BurstSR is proposed in this paper. The main components of the BSRT include the Pyramid Flow-based Deformable alignment module (Pyramid FG-DCN) and the Swin Transformer-based backbone. Compared with the previous methods, the proposed Pyramid FG-DCN can greatly improve the alignment performance and alleviate the effect of noises. Meanwhile, Swin Transformer blocks and groups in our backbone can make more effective use of global contextual information in multi-frames and further improve the performance through the self-attention mechanism. Our results on both synthetic and real-world datasets demonstrate that our method achieves a state-of-the-art performance and recovers more plausible and pleasing visual results. Furthermore, our proposed BSRT wins 1st place in real-world track of the NTIRE 2022 Burst Super-Resolution Challenge.

\section*{Acknowledgment} This work was supported by the National Natural Science Foundation of China (NSFC) under grants No.61872067.

{\small
\bibliographystyle{ieee_fullname}
\bibliography{main}
}

\end{document}